\pdfoutput=1

\documentclass[letterpaper, 10 pt, conference]{ieeeconf}  % Comment this line out if you need a4paper

\usepackage{amssymb}
\usepackage{amsfonts}
\usepackage{multirow}
\usepackage{textcomp}
\usepackage{bbding}
\usepackage{setspace}
\usepackage{amsmath}
\usepackage{xfrac}    % Works better with other fonts
\usepackage{algpseudocode}
\usepackage{lipsum} % Dummy text package
\usepackage{wrapfig}
\usepackage{multicol}
\usepackage[ruled,vlined]{algorithm2e}
\usepackage{graphicx}
\usepackage{booktabs}
\usepackage{tabularx} 
\usepackage{pifont}
\usepackage{bm}
\usepackage{dsfont}

\IEEEoverridecommandlockouts                              % This command is only needed if 
\usepackage{kotex}
\usepackage{color}

\usepackage{url}

\title{\LARGE \bf
RE-TRIP : Reflectivity Instance Augmented Triangle Descriptor for 3D Place Recognition}

\author{Yechan Park$^{1}$, Gyuhyeon Pak$^{2}$ and Euntai Kim$^{2,*}$  % <-this % stops a space
\thanks{This work was supported by Korea Evaluation Institute Of Industrial Technology (KEIT) grant funded by the Korea government(MOTIE) (No.20023455, Development of Cooperate Mapping, Environment Recognition and Autonomous Driving Technology for Multi Mobile Robots Operating in Large-scale Indoor Workspace). This work was also supported by Institute of Information \& communications Technology Planning \& Evaluation (IITP) grant funded by the Korea government(MSIT) (No.2022-0-01025, Development of core technology for mobile manipulator for 5G edge-based transportation and manipulation)}
\thanks{*is that the corresponding author}% <-this % stops a space
\thanks{$^{1}$Yechan Park is with the Department of Vehicle Convergence Engineering, Yonsei University, Seoul 03722, South Korea
        {\tt\small pyc5714@yonsei.ac.kr}}%
\thanks{$^{2}$Gyuhyeon Pak and Euntai Kim are with the Department of Electrical and Electronic Engineering, Yonsei University, Seoul 03722, South Korea {\tt\small gh.pak@yonsei.ac.kr, etkim@yonsei.ac.kr}.}%
}

\begin{document}

\maketitle
\thispagestyle{empty}
\pagestyle{empty}

\begin{abstract}
While most people associate LiDAR primarily with its ability to measure distances and provide geometric information about the environment (via point clouds), LiDAR also captures additional data, including reflectivity or intensity values. Unfortunately, when LiDAR is applied to Place Recognition (PR) in mobile robotics, most previous works on LiDAR-based PR rely only on geometric measurements, neglecting the additional reflectivity information that LiDAR provides.
In this paper, we propose a novel descriptor for 3D PR, named RE-TRIP (REflectivity-instance augmented TRIangle descriPtor). This new descriptor leverages both geometric measurements and reflectivity to enhance robustness in challenging scenarios such as geometric degeneracy, high geometric similarity, and the presence of dynamic objects. To implement RE-TRIP in real-world applications, we further propose (1) keypoint extraction method, (2) key instance segmentation method, (3) RE-TRIP matching method, and (4) reflectivity combined loop verification method.
Finally, we conduct a series of experiments to demonstrate the effectiveness of RE-TRIP. Applied to public datasets (i.e., HELIPR, FusionPortable) containing diverse scenarios—including long corridors, bridges, large-scale urban areas, and highly dynamic environments—our experimental results show that the proposed method outperforms existing state-of-the-art methods in terms of Scan Context, Intensity Scan Context and STD. Our code is available at : \url{https://github.com/pyc5714/RE-TRIP}.
\end{abstract}

\section{INTRODUCTION}  
Place recognition enables a robot to identify previously visited locations despite changes in viewpoint or time. PR plays a crucial role in various mobile robotics applications, such as Simultaneous Localization and Mapping (SLAM) \cite{shan2020lio,mur2015orb}, loop closure \cite{zou2024lta,wang2021intensity}, re-localization \cite{kim20191}, and initial pose estimation \cite{koide2024tightly}. Traditionally, visual sensors have been primarily used to solve this problem \cite{bay2008speeded,lowry2015visual,rublee2011orb,galvez2012bags}. However, visual sensor-based methods often suffer from inherent issues due to significant variations in viewpoint and illumination.

As an alternative to visual PR, LiDAR sensors are gaining attention because of their robustness to viewpoint changes and illumination variations \cite{steder2010robust,rusu2010fast,gupta2024effectively}. Several studies \cite{kim2021scan,he2016m2dp,yuan2024btc,yuan2023std,jiang2019triangle} have focused on utilizing LiDAR's accurate geometric measurements to capture the structural information of environments, proposing effective descriptors \cite{kim2021scan,he2016m2dp} to represent structural appearance. Some approaches \cite{yuan2024btc,yuan2023std,jiang2019triangle} extract geometric keypoints and connect them into various polygonal shapes to achieve translation and rotation invariance. While these methods maintain robustness and distinctiveness in environments with well-defined geometric structures, they struggle in geometrically ambiguous and dynamic environments, i.e., (i) geometric degeneracy, (ii) high geometric similarity, or (iii) constantly changing geometry due to dynamic objects. These challenges are amplified in long trajectories, as the larger database increases ambiguity, making place recognition more difficult.

%---------------------------------------
%              Fig 1
%---------------------------------------
\begin{figure}[t] % 't' option forces the figure to the top of the column
    \centering
    \includegraphics[width=\linewidth]{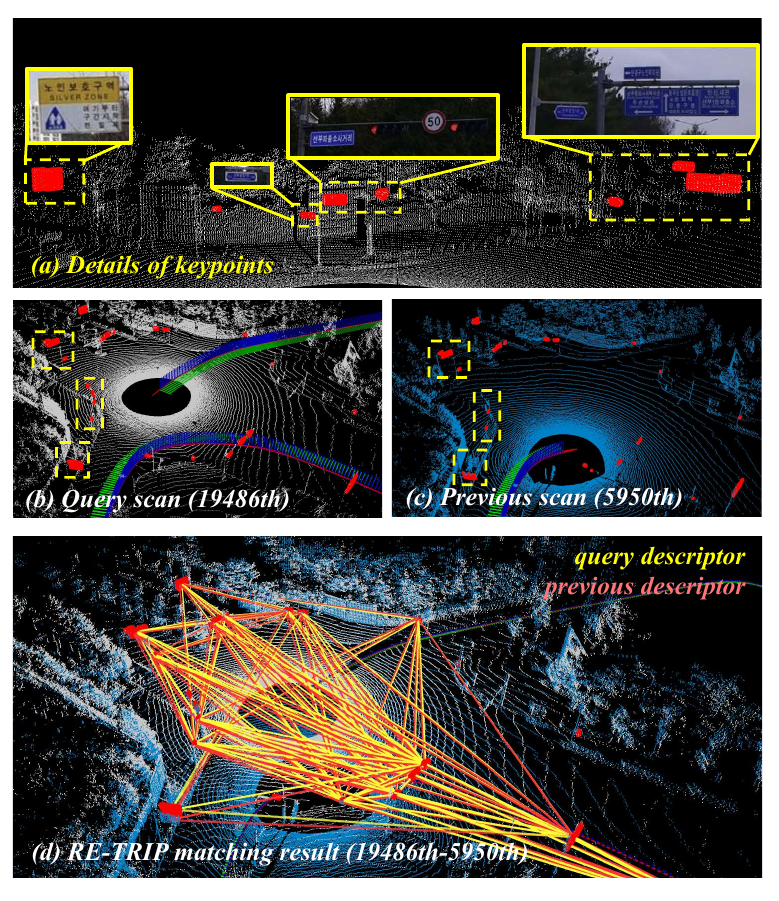}
    \vspace{-20pt}
    \caption{Overview of the proposed method.
(a) Keypoints (red points) extracted from the high-reflectivity objects (yellow box). (b) LiDAR scan when revisiting the same location. (c) LiDAR scan from the previous visit. Despite (b) and (c) are captured from different viewpoints due to opposite driving directions, keypoints (red points) are consistently detected from the high-reflectivity objects (yellow box).
(d) By associating these distinctive instances into a triangular shape, RE-TRIP effectively retrieves the same location from the database.}
    \label{fig:fig1}
    \vspace{-20pt}
\end{figure}
%---------------------------------------
%              Fig 1
%---------------------------------------

To address these issues, recent studies \cite{pfreundschuh2024coin,du2023real,zhao2024fmcw,zhang2023ri} have integrated additional LiDAR information such as intensity, reflectivity, and velocity. In particular, \cite{pfreundschuh2024coin,du2023real,wang2020intensity} leverage LiDAR’s intensity measurements to capture texture information and mitigate challenges related to geometric degeneracy. However, these approaches commonly encode LiDAR intensity as 2D images or 2.5D matrices, which compromises geometric structure information during the process.

In this paper we propose a novel reflectivity and geometry combined place recognition framework, termed REflectivity instance augmented TRIangle descriPtor for 3d place recognition (RE-TRIP), to address the limitations of existing geometry-based and reflectivity-based methods. Our approach leverages object reflectivity properties to consistently detect distinctive reflectivity instances at the same locations as shown in Fig \ref{fig:fig1}(a). To recognize the structural associations of instances, we adopt the triangle descriptor, inspired by STD \cite{yuan2023std}, as the triangular shape is inherently invariant to translation and rotation. Finally, we evaluate the similarity between the reflectivity instances and the triangular shapes to determine whether the location has been previously visited, as shown in Fig. \ref{fig:fig1}(d).
The main contributions of this work are as follows:

\begin{itemize}
\item
We propose the novel Reflectivity Augmented Triangle Descriptor (RE-TRIP) for 3D place recognition, which focuses on leveraging the object reflectivity property and their structural associations.
\item 
We develop a simple and robust key instance segmentation method based on two type of keypoints: Absolute Reflectivity Points (ARP) and Relative Reflectivity Points (RRP). 
\item
We introduce a RE-TRIP matching and reflectivity-combined geometric verification approach to fully exploit the distinctiveness of key instances and reflectivity layers.
\item
We validate our approach through experiments in various scenarios, demonstrating its effectiveness in geometrically ambiguous and dynamic environments. To support the wider robotics community, we have made our algorithm open-source on \url{https://github.com/pyc5714/RE-TRIP}.
\end{itemize}

\section{RELATED WORK}

Place recognition is a critical task in robot navigation and has been extensively studied. Existing methods can be broadly categorized into local descriptor-based approaches, global descriptor-based approaches, and intensity-assisted methods.

\subsection{Local Descriptor-Based Place Recognition}

Local descriptors aim to capture fine-grained geometric details for discriminative matching. In visual place recognition, descriptors like SIFT \cite{lowe2004distinctive}, ORB \cite{rublee2011orb}, and BRIEF \cite{calonder2010brief} are widely used. Extending these to 3D point clouds, methods such as 3D SIFT \cite{scovanner20073}, SHOT \cite{salti2014shot}, and 3D-SURF \cite{knopp2010hough} extract local features from 3D LiDAR. While effective in capturing detailed geometry, these methods are sensitive to point cloud density, resolution, and viewpoint changes, limiting their robustness in diverse environments.

\subsection{Global Descriptor-Based Place Recognition}

To mitigate the limitations of local descriptors, global descriptors summarize the overall structural information of the environment, enhancing robustness to local noise and density variations. Methods like M2DP \cite{he2016m2dp}, Scan Context (SC) \cite{kim2018scan}, and SC++ \cite{kim2021scan} project 3D point clouds onto 2D planes to encode the surrounding environment's structural appearance. Although projection-based global descriptors are more robust to local variations, they struggle with translation and viewpoint changes.

To improve invariance, some methods extract keypoints from projection planes and leverage their spatial relationships. For example, Gupta et al. \cite{gupta2024effectively} extract ORB features \cite{rublee2011orb} from bird's eye view images and match them using a binary tree database. STD \cite{yuan2023std} and BTC \cite{yuan2024btc} project point clouds onto reference planes, extract keypoints, and construct triangle descriptors. By indirectly using projection planes, these methods enhance translation and rotation invariance. However, they still face challenges in geometrically ambiguous and dynamic environments.
 
\subsection{Intensity-Assisted Methods}
To address the above challenges, some studies \cite{pfreundschuh2024coin,zhang2023ri,du2023real} incorporate texture features from LiDAR intensity measurements. For instance, ISC \cite{wang2020intensity} integrates intensity into subspaces to compute scan similarity. Other works \cite{di2021visual, shan2021robust} apply visual place recognition techniques to 2D intensity images extracted from 3D LiDAR. While intensity-assisted approaches provide additional texture information, they still suffer from the fundamental limitations of local and global descriptors, including sensitivity to noise and susceptibility to rotation and translation variations.

We propose the RE-TRIP framework, which integrates reflectivity information—a refined LiDAR measurement that accounts for distance and angle of incidence—with geometric structure. By leveraging reflectivity-based key instances and geometric triangle descriptors, our method provides more consistent and distinctive place recognition. This approach addresses the shortcomings of previous methods, enhancing robustness against geometrically ambiguous and dynamic environments.

\begin{figure*}[t] 
    \centering
    \includegraphics[width=\textwidth]{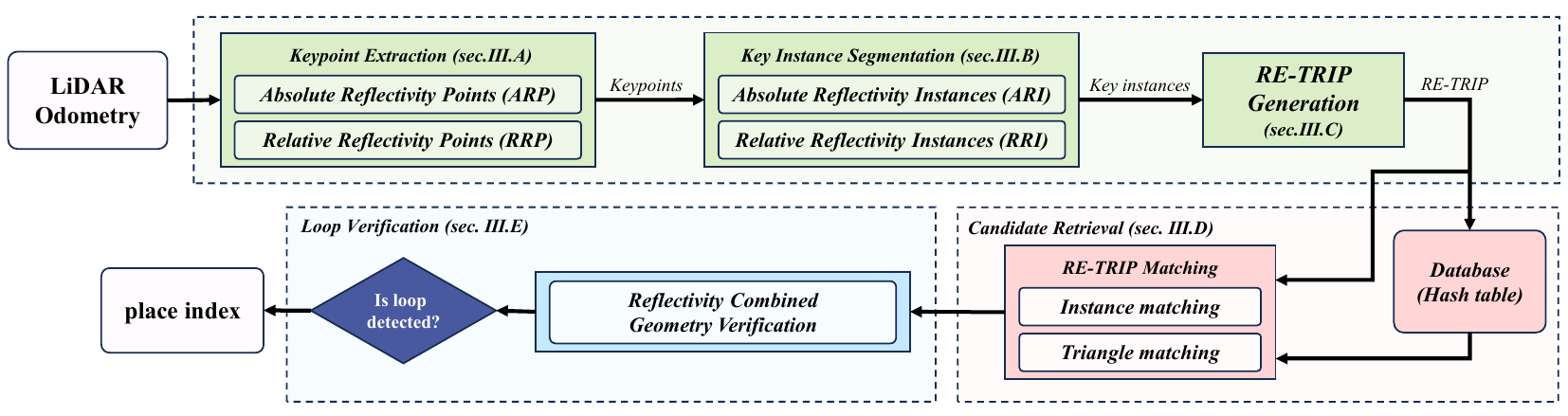}
    \vspace{-20pt}
    \caption{The overall framework of our proposed method consists of four steps. First, keypoints are extracted based on reflectivity measurement, identifying Absolute Reflectivity Points (ARP) and Relative Reflectivity Points (RRP). Then keypoints are grouped into instances and a small number of significant key instances are selected to maintain efficiency while keeping computational costs low. Next, we generate Reflectivity Instance Augmented Triangle Descriptors (RE-TRIP), which incorporate both geometric and reflectivity information. RE-TRIPs are then retrieved from the hash table and matched based on key instance similarity and side lengths. Finally, candidate frames are verified through reflectivity-combined geometric verification.}
    \label{fig:overview}
    \vspace{-10pt}
\end{figure*}

\section{METHODOLOGY}
 Fig. \ref{fig:overview} shows the overall framework of our proposed method. All the steps in the figure are explained in the subsequent sections. 

\subsection{Keypoint Extraction} 
\label{sec:Keypoint Extraction}
Let us suppose that point cloud $\mathcal{P} = \{\mathbf{p}_1,\mathbf{p}_2, \cdot \cdot \cdot ,\mathbf{p}_N\}$ is given from a LiDAR scan, where $\mathbf{p}$ = [$\mathit{x,y,z,r}$]. Here, $\mathit{x}$, $\mathit{y}$, and $\mathit{z}$ are a Cartesian coordinate and $\mathit{r}$ is the reflectivity measured by the LiDAR. We decompose the point cloud $\mathcal{P}$ into three exclusive categories: Absolute Reflectivity Points (ARP), Relative Reflectivity Points (RRP) and the remaining points. We will denote ARP, RRP, and the remaining points by  $\mathcal{P^A}$, $\mathcal{P^R}$ and  $(\mathcal{P^S})'$, respectively. If we define $\mathcal{P^S} = \mathcal{P^A} \cup \mathcal{P^R}$, $\mathcal{P} = \mathcal{P^S} \cup (\mathcal{P^S})'$ and $ \mathcal{P^S} \cap (\mathcal{P^S})' = \varnothing $.

\subsubsection{\textbf{Absolute Reflectivity Point}}
In real-world environments, various objects are coated with reflective materials to ensure high visibility. Fig. \ref{fig:fig1}(b) and (c) show the detection results of points with high reflectivity $r$. These objects can be observed regardless of density or viewpoint because they return a significant portion of the light emitted from the LiDAR emitter. Additionally, even if part of the object is occluded by other dynamic objects, the remaining high-reflectivity regions allow for consistent detection of the object. We refer to the points as Absolute Reflectivity Points (ARP) and they are defined by
\begin{equation}
    \mathcal{P^A} = \{ \mathbf{p}_i=(\mathit{x_i,y_i,z_i,r_i}) \mid \mathbf{p}_i \in \mathcal{P}, \ \frac{\mathit{r_i}-\mu_{\mathit{r}}}{\sigma_{\mathit{r}}} > z_A \},
\end{equation}
where 
\begin{equation} \label{eq:reflectivity mean and sttdev}
\mu_{\mathit{r}}=\frac{1}{|\mathcal{P}|}\sum_{\mathit{r_i} \in \mathcal{P}}\mathit{r_i} , \ \sigma_{\mathit{r}}^2=\frac{1}{|\mathcal{P}|}\sum_{\mathit{r_i} \in \mathcal{P}}\left(\mathit{r_i}-\mu_{\mathit{r}}\right)^2.
\end{equation}

\noindent In other words, ARP are the points whose reflectivity is higher than the mean reflectivity $\mu_{\mathit{r}}$ by more than $z_A \sigma_r$. 

\subsubsection{\textbf{Relative Reflectivity Point}}
Points whose reflectivity significantly differs from that of neighboring points can also be consistently detected, regardless of variations in viewing conditions. We refer to these points as Relative Reflectivity Points (RRP) and they are defined by
\begin{equation}
    \label{equ:P_R}
    \mathcal{P^R} = \{ \mathbf{p}_i=(\mathit{x_i,y_i,z_i,r_i}) \mid  \ \Delta \mathit{r_i} >  \delta_r \},
\end{equation}
where
\begin{equation} 
\Delta \mathit{r_i} = \frac{1}{|\mathcal{S}_i|} \sum_{j:\mathbf{p}_j \in \mathcal{S}_i} \left(\mathit{r_i} - \mathit{r_j}\right)^2 ,
\end{equation}
where $\mathcal{S}_i$ is the set of consecutive points around $\mathbf{p}_i$ and $\delta_r$ is a pre-defined threshold. 

\subsection{Key Instance Segmentation} \label{sec:key instance segmentation} 
After extracting keypoints $\mathcal{P}^A$ and $\mathcal{P}^R$, the keypoint set is refined through key instance segmentation, which groups keypoints into distinct clusters. This process effectively suppresses noisy keypoints by eliminating scattered outliers. The approach proves effective as most noisy points fail to form meaningful clusters, whereas the proposed keypoints are designed to represent specific objects. Managing keypoints at the instance level contributes to a more robust and reliable descriptor generation process.

\subsubsection{\textbf{Keypoint Clustering}}
Using an Euclidean distance based point cloud clustering function $\mathsf{EuclideanCluster}$ $\mathsf{Extraction}$ $\mathsf{(EC)}$  from PCL \cite{rusu20113d}, we segment two sets of keypoints  $\mathcal{P^A}$ and $\mathcal{P^R}$ into two sets of clusters $\mathcal{C^A}$ and $\mathcal{C^R}$ by
\begin{equation}
    \mathcal{C}^\mathcal{X} =\{\mathsf{EC.extract}(\mathcal{P}^\mathcal{X})\},(\mathcal{X}=\mathcal{A},\mathcal{R})
\end{equation}
where the cluster set $\mathcal{C}^\mathcal{X}=\{\mathbf{C}^\mathcal{X}_{1},\mathbf{C}^\mathcal{X}_{2},\dots,\mathbf{C}^\mathcal{X}_{|{\mathcal{C}^\mathcal{X}}|}\}$ and each cluster $\mathbf{C}^\mathcal{X}_{i}=$ $\{\mathbf{p}^\mathcal{X}_{1},$$\mathbf{p}^\mathcal{X}_2,$ $\dots$$,\mathbf{p}^\mathcal{X}_{|\bm{\mathcal{C}}_{i}^{\mathcal{X}}|}\}$.

\subsubsection{\textbf{Instance Extraction}}
Then, we redefine each cluster $\mathbf{C}^\mathcal{X}_{i}, (\mathcal{X}=\mathcal{A},\mathcal{R})$, a reflectivity instance 
\begin{equation}
    \bm{\mathcal{I}}^\mathcal{X}_i = \{ \bm{\mu}^\mathcal{X}_i, \mathbb{I}(\mathbf{C}^\mathcal{X}_{i}), |\mathbf{C}^\mathcal{X}_{i}| \}, 
\end{equation}
\noindent where $\bm{\mu}^\mathcal{X}_i=\frac{1}{|\mathbf{C}^\mathcal{X}_{i}|}\sum\limits_{\mathbf{p}_j \in \mathbf{C}^\mathcal{X}_{i}} \mathbf{p}_j$; $\mathbb{I}(\cdot)$ is an indicator function which returns 1 if the corresponding cluster $\mathbf{C}^\mathcal{X}_{i}$ comes from ARP $\mathcal{P^A}$ but returns 0 if the cluster comes from RRP $\mathcal{P^R}$; $|\mathbf{C}^\mathcal{X}_{i}|$ denotes the size of the cluster.

This transformation is crucial for the subsequent step, where we construct triangular descriptors. Instead of handling a large set of individual points, the instance representation allows for efficient and structured processing while retaining the key characteristics necessary for descriptor generation and comparison.

Finally, we build two sets of instances $\mathcal{I^\mathcal{A}}=\{\bm{\mathcal{I}}^{\mathcal{A}}_{1},\bm{\mathcal{I}}^{\mathcal{A}}_{2},\dots,$ $\bm{\mathcal{I}}^{\mathcal{A}}_{|\mathcal{I^A}|} \}$ and $\mathcal{I^R}=\{\bm{\mathcal{I}}^{\mathcal{R}}_{1}, \bm{\mathcal{I}}^{{\mathcal{R}}}_{2}, \dots, \bm{\mathcal{I}}^{{\mathcal{R}}}_{|\mathcal{I^R}|} \}$, referred to as ARI (Absolute Reflectivity Instances) and RRI (Relative Reflectivity Instances), respectively. Obviously,  $|\mathcal{I^A}| = |\mathcal{C^A}|$ and  $|\mathcal{I^R}| = |\mathcal{C^R}|$.

\subsubsection{\textbf{Key Instance Set Building}}
To enhance both efficiency and reliability, we then combine these two sets of instances, $\mathcal{I}^\mathcal{A}$ and $\mathcal{I}^\mathcal{R}$, to form a comprehensive key instance set $\mathcal{S}$. The key instance set $\mathcal{S}$ consists of the top $k$ instances based on instance size, as larger instances imply higher detection reliability. Specifically, we prioritize ARI over RRI because ARI instances, derived from consistently high reflectivity points, offer more reliable detection across varying conditions. The selection of $\mathcal{S}$ is performed as follow, if \( |\mathcal{I}^{\mathcal{A}}| \geq k \), we select the top \( k \) instances from \( \mathcal{I}^{\mathcal{A}} \), i.e., 
\[
\mathcal{S} = \{ \bm{\mathcal{I}}^{\mathcal{A}}_{1}, \bm{\mathcal{I}}^{\mathcal{A}}_{2}, \dots, \bm{\mathcal{I}}^{\mathcal{A}}_{k} \}.
\]
However, if \( |\mathcal{I}^{\mathcal{A}}| < k \), all instances from \( {\mathcal{I}}^{\mathcal{A}} \) are selected, and the remaining \( k - |{\mathcal{I}}^{\mathcal{A}}| \) instances are filled from \( {\mathcal{I}}^{\mathcal{R}} \), i.e., 
\[
\mathcal{S} = \{ \bm{\mathcal{I}}^{\mathcal{A}}_{1}, \bm{\mathcal{I}}^{\mathcal{A}}_{2}, \dots, \bm{\mathcal{I}}^{\mathcal{A}}_{|\mathcal{I}^{\mathcal{A}}|} \} \cup \{ \bm{\mathcal{I}}^{\mathcal{R}}_{1}, \bm{\mathcal{I}}^{\mathcal{R}}_{2}, \dots, \bm{\mathcal{I}}^{\mathcal{R}}_{k - |\mathcal{I}^{\mathcal{R}}|} \}.
\]
By selecting key instances based on both reliability and instance size, we can represent the places with a small number of key instances (around 20 instances) while maintaining consistency.

\subsection{Reflectivity Instance Augmented Triangle Descriptor}
After constructing the key instance set $\mathcal{S}$, we derive structural associations by connecting the centroids (\( \bm{\mu} \)) of three instances from $\mathcal{S}$, forming a triangle descriptor. The descriptor, referred to as RE-TRIP, is denoted as:
\begin{equation}
    \bm{\mathcal{D}} = \{\bm{\mathcal{I}}_1,\bm{\mathcal{I}}_2,\bm{\mathcal{I}}_3,l_{12}, l_{23}, l_{13},\mathbf{q},t\},
\end{equation}
where \( \bm{\mathcal{I}}_1, \bm{\mathcal{I}}_2, \bm{\mathcal{I}}_3 \) represent the three instances  that constitute the vertices of the triangle; \( l_{12}, l_{23}, l_{13} \) are three sides of the triangle; \( \mathbf{q} \) is the centroid of the triangle; and \( t \) represents the frame index where the descriptor was generated.

While the triangle descriptor \cite{yuan2023std} captures geometric relationships, relying solely on three side lengths for retrieval may lead to local ambiguity \cite{yin2024outram}. To address the problem, additional attributes are required to enhance distinctiveness. RE-TRIP incorporates two key attributes: the geometric structure of the triangle and the reflectivity instances of its vertices. By integrating reflectivity information, RE-TRIP not only captures spatial relationships between points but also encodes material characteristics, making the descriptor more distinctive and reliable.

\begin{figure}[t]
    \centering
    \includegraphics[width=\linewidth]{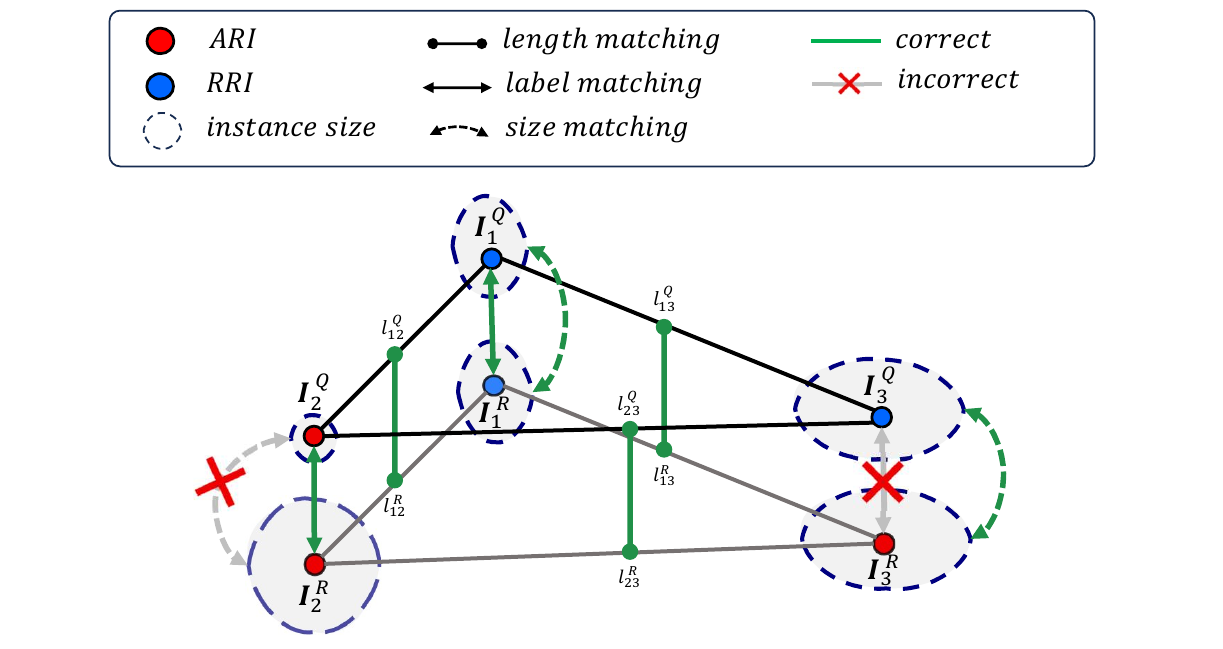}
    \caption{Description of RE-TRIP matching between $\bm{\mathcal{D}}^Q$ and $\bm{\mathcal{D}}^R$. It consists of three steps: (1) side length matching, (2) label matching, and (3) instance size comparison. }
    \label{fig:RAM}
    \vspace{-15pt}
\end{figure}

\subsection{Candidate Retrieval} \label{sec:candidate retrieval} 
To leverage augmented descriptors for retrieval, the current set of RE-TRIPs, denoted as  $\mathcal{D}^Q = \{\bm{\mathcal{D}}^Q_i\}^{N}_{i=1}$, is stored in a hash table. The sorted lengths of the triangle sides are used as the hash key, following a similar approach to that used in STD \cite{yuan2023std}.  

To mitigate aforementioned ambiguity, we employ two strategies: using larger triangles and instance matching. Small triangles with the same shape tend to appear repeatedly in different locations. In contrast, larger triangles are less prone to being repeated in different locations, enhancing distinctiveness. However, constructing larger triangles requires connecting each vertex with a greater number of surrounding vertices, leading to a significant increase in both the number of vertices and the total number of triangles.

Fortunately, as described in Section \ref{sec:key instance segmentation}, key instance segmentation reduces the number of vertices to around 20, making larger triangle construction feasible without excessive computational cost. Furthermore, as illustrated in Fig. \ref{fig:RAM}, combining instance matching with side length matching significantly enhances distinctiveness, even in geometrically similar environments. After completing the matching of all $N$ descriptors in the current set $\mathcal{D}^Q$, we analyze the retrieved descriptors, denoted as $\mathcal{D}^R = \{\bm{\mathcal{D}}^{R}_{i}\}^{M}_{i=1}$, to determine the final loop frame candidates.

\subsection{Loop Verification}
For false positive rejection, we propose a reflectivity-combined geometry verification method. We reference STD \cite{yuan2023std} for calculating plane voxels by analyzing the covariance matrix of each voxel, denoted as $\bm{\mathit{v}}=\{\mathbf{p}_1$,$\mathbf{p}_2$, $\dots,\mathbf{p}_N\}$ in the current point cloud. We then calculate the mean reflectivity of each plane as $\mu_r^v=\frac{1}{|\bm{\mathit{v}}|}\sum_{\mathit{r}_i \in {\mathit{v}}} \mathit{r}_i$, dividing the reflectivity layer. The reflectivity layer is categorized into five levels based on how many times $\mu_r^v$ exceeds $\mu_r$ by $\lambda\sigma_r\mathit{z}_L$, where $\sigma_r$ and $\mu_r$ are computed from Eq. \ref{eq:reflectivity mean and sttdev}, $\mathit{z}_L$ is the layer threshold, and $\lambda$ represents the layer index.

Next, the matching descriptors $\mathcal{D}^R$ are used to estimate the transformation $\mathbf{T}_Q^R=\left(\mathbf{R}_Q^R, \mathbf{t}_Q^R\right) \in SE(3)$ between the current and reference frames using Singular Value Decomposition (SVD). With $\mathbf{T}_Q^R$, we compute the overlap between the planes in the current frame ($\pi^Q=\left[\left(\mathbf{q}_\lambda^Q, \mathbf{u}_1^Q, \lambda_1^Q\right), \ldots, \left(\mathbf{q}_n^Q, \mathbf{u}_n^Q, \lambda_n^Q\right)\right]$) and the retrieved frame ($\pi^R=\left[\left(\mathbf{q}_1^R, \mathbf{u}_1^R, \lambda_1^R\right), \ldots, \left(\mathbf{q}_m^R, \mathbf{u}_m^R, \lambda_m^R\right)\right]$), where $\mathbf{q}$ denotes the center point, $\mathbf{u}$ is the normal vector, and $\lambda$ is the layer index of each plane. The number of planes in the current and retrieved frames are denoted by $n$ and $m$, respectively.

To compute plane overlap, we use the formula provided in STD \cite{yuan2023std}:
\begin{equation}
\left\|\mathbf{R}_Q^R \mathbf{u}_i^Q - \mathbf{u}_j^R\right\|_2 < \sigma_n,
\end{equation}
\begin{equation}
\left(\mathbf{u}_j^R\right)^T \left(\mathbf{T}_Q^R \mathbf{q}_i^Q - \mathbf{q}_j^R\right) < \sigma_d,
\end{equation}
where $\sigma_n$ and $\sigma_d$ are pre-set hyperparameters.
%and $\mathbf{q}_j$ is the center point of the nearest plane in the retrieved frame.

%---------------------------------------
%             fig : PR curve
%---------------------------------------

\begin{figure*}[t] % 't' option forces the figure to the top of the column
    \centering
    \includegraphics[width=\linewidth]{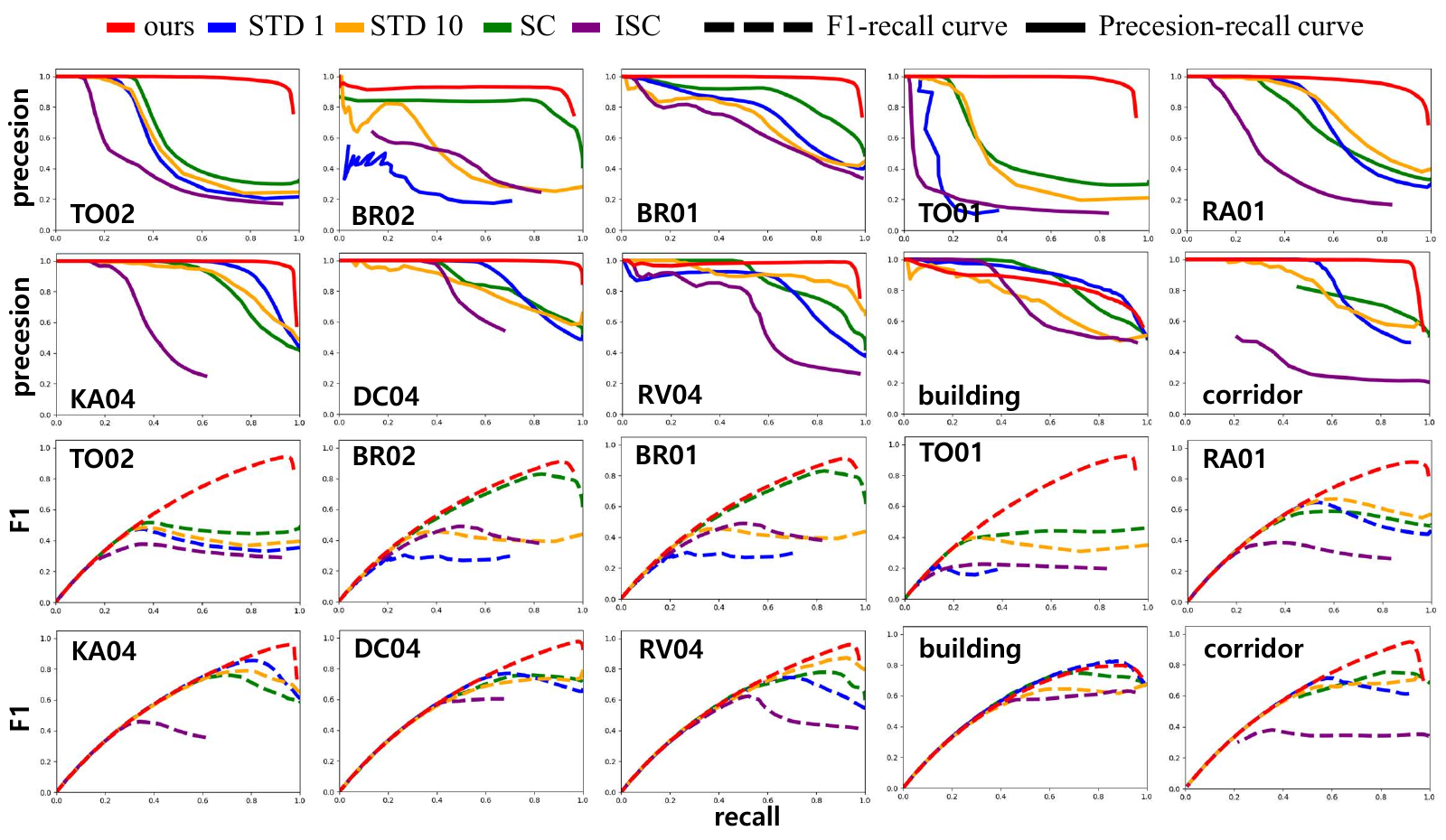}
    \vspace{-20pt}
    \caption{Precision-Recall curves (top two rows) and F1 Score-Recall curves (bottom two rows) for HELIPR and FusionPortable.}
    \label{fig:PRcurve and FRcurve}
    \vspace{-5pt}
\end{figure*}

%---------------------------------------
%             fig : PR curve
%---------------------------------------

To enhance accuracy in environments with high spatial similarity, we introduce layer matching based on the layer index:
\begin{equation}
\left|\lambda_i - \lambda_j\right| < \sigma_\lambda,   
\end{equation}
where $\sigma_{\lambda}$ is a pre-set threshold and $\lambda_j$ is the nearest plane's layer to $\lambda_i$. Finally, we calculate the percentage of plane coincidence, which is used to eliminate false positives.

\section{EXPERIMENTS}

\subsection{Dataset}
We evaluate the proposed method on two public datasets: HELIPR \cite{jung2023helipr}, an extended version of MULRAN \cite{kim2020mulran}, and FusionPortable \cite{jiao2022fusionportable}. These datasets are selected for their various scenarios, including geometric degeneracy, high geometric similarity, and dynamic objects. Both datasets were recorded using an Ouster LiDAR, which provides direct reflectivity measurements, making them well-suited for our approach. HELIPR includes various long-trajectory outdoor sequences from TOWN, BRIDGE, ROUNDABOUT, KAIST, DCC, and RIVERSIDE, abbreviated as TO, BR, RA, KA, DC, and RV. For indoor environment experiments, we also experimented the building and corridor sequences from FusionPortable.

\subsection{Experimental Setup}
All experiments are implemented in C++ and ROS1 on a PC equipped with an Intel i9-10920 processor and 128GB of RAM. 

For comparison, we evaluate the following baselines: Scan Context (SC) \cite{kim2018scan}, Intensity Scan Context (ISC) \cite{wang2020intensity}, and STD \cite{yuan2023std}. We implemented these baselines using the default settings from the respective papers. However, to ensure a fair comparison with our approach, we made a few adjustment. Specifically, for ISC \cite{wang2020intensity}, since it also utilizes intensity measurements, we use the same reflectivity data provided by the LiDAR to compute ISC similarity. Additionally, we evaluate STD \cite{yuan2023std} in two configurations: STD-1 and STD-10. Similar to other methods, STD-1 performs loop detection on every frame, while STD-10 follows the approach described in STD \cite{yuan2023std}, conducting loop detection only on keyframes that accumulate points from 10 consecutive frames. However, since the FusionPortable indoor datasets are acquired using a handheld device, drift occurs in the accumulated points between adjacent frames. As a result, we experiment only with STD-1 on the FusionPortable.

%---------------------------------------
%              Table 3 : max F1 score and AUC
%---------------------------------------
\begin{table}[!t]
\centering
\caption{Comparison of Max F1 Score and AUC for State-of-the-Art Place Recognition Methods}
\label{table:max F1 and AUC}
\resizebox{0.49\textwidth}{!}{
\begin{tabular}{c|cc|cc|cc|cc|cc}
\toprule
Dataset & \multicolumn{2}{c}{Ours} & \multicolumn{2}{c}{STD-10} & \multicolumn{2}{c}{STD-1} & \multicolumn{2}{c}{Scan context} & \multicolumn{2}{c}{ISC} \\ 
  & AUC & F1 & AUC & F1 & AUC & F1 & AUC & F1 & AUC & F1 \\ 
 \midrule
TO02 & \textbf{0.96} & \textbf{0.94} & 0.56 & 0.49 & 0.53 & 0.47 & \underline{0.62} & \underline{0.52} & 0.39 & 0.38 \\ 
BR02 & \textbf{0.89} & \textbf{0.91} & \underline{0.48} & 0.46 & 0.20 & 0.30 & 0.41 & \underline{0.83} & 0.41 & 0.49 \\ 
BR01 & \textbf{0.98} & \textbf{0.95} & 0.70 & 0.64 & 0.76 & 0.68 & \underline{0.87} & \underline{0.78} & 0.65 & 0.60 \\ 
TO01 & \textbf{0.94} & \textbf{0.92} & 0.46 & 0.40 & 0.10 & 0.22 & \underline{0.52} & \underline{0.48} & 0.17 & 0.23 \\ 
RA01 & \textbf{0.96} & \textbf{0.91} & \underline{0.78} & \underline{0.67} & 0.72 & 0.65 & 0.70 & 0.59 & 0.38 & 0.38 \\ 
KA04 & \textbf{0.99} & \textbf{0.97} & 0.88 & 0.79 & \underline{0.93} & \underline{0.86} & 0.86 & 0.76 & 0.41 & 0.46 \\ 
DC04 & \textbf{0.99} & \textbf{0.98} & 0.82 & \underline{0.79} & \underline{0.88} & 0.77 & 0.86 & 0.76 & 0.59 & 0.60 \\ 
RV04 & \textbf{0.95} & \textbf{0.96} & \underline{0.90} & \underline{0.87} & 0.80 & 0.75 & 0.88 & 0.78 & 0.63 & 0.62 \\ 
 \midrule
 building & 0.84 & \underline{0.80} & N/A & N/A & \textbf{0.89} & \textbf{0.82} & \underline{0.86} & 0.75 & 0.71 & 0.63 \\ 
corridor & \textbf{0.95} & \textbf{0.95} & N/A & N/A & \underline{0.78} & 0.71 & 0.38 & \underline{0.75} & 0.22 & 0.38 \\ 
\bottomrule
\end{tabular}
}
\vspace{-10pt}
\end{table}

%---------------------------------------
%              Table 3 : max F1 score and AUC
%---------------------------------------

%---------------------------------------
%             fig : candidate_ablation
%---------------------------------------
\begin{figure*}[t] % 't' option forces the figure to the top of the column
    \centering
    \includegraphics[width=\linewidth]{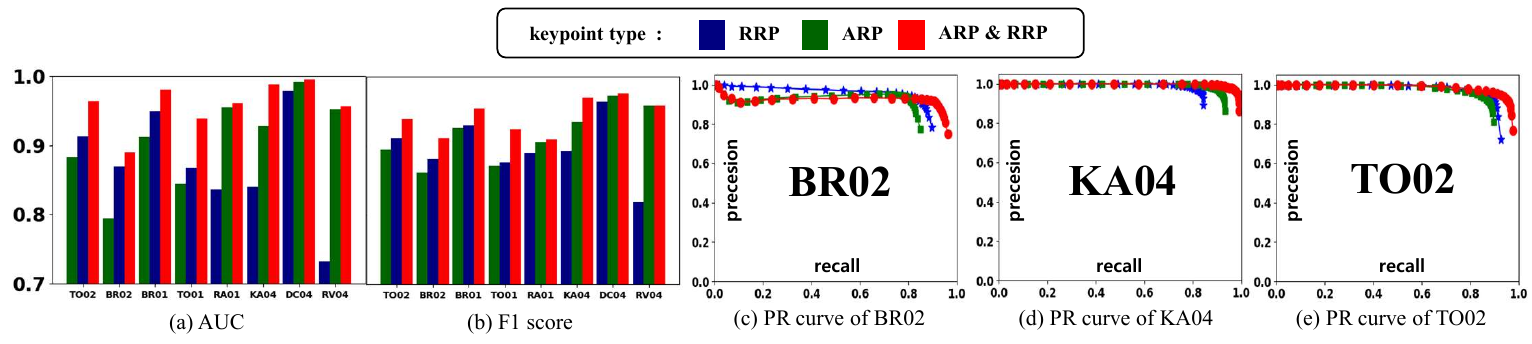}
    \vspace{-20pt}
    \caption{Keypoint ablation results. (a) and (b) compare the performance of ARP, RRP, and the combined approach across all sequences of HELIPR. The combined use of both ARP and RRP consistently yields superior results, demonstrating the complementary nature of both keypoints.}
    \label{fig:keypoint ablation}
    \vspace{-10pt}
\end{figure*}
%---------------------------------------
%             fig : candidate_ablation
%---------------------------------------
The ground truth criteria, which are uniformly applied to all methods, are determined separately for outdoor (HELIPR) and indoor (FusionPortable) datasets. We define successful place recognition as occurring when the current place and the candidate place are within 20m in outdoor environments and within 4m for indoor environments. To avoid redundant scans, we perform the evaluation every 0.25m outdoors and every 0.1m indoors.

To evaluate the proposed method, we use consistent settings: 20 key instances, 10 loop candidates, and plane overlap thresholds of $\sigma_n = 0.2$, $\sigma_d = 0.3$, and $\sigma_\lambda = 3$. The ARP threshold $z_A$ is set to 3.5 for indoor and 4.5 for outdoor environments. Our method uses 10 loop candidates, while STD is set to 50. A performance comparison based on loop candidates is presented in the Table \ref{table:loop candidates}.

\begin{table}[!t]
\centering
\caption{Ablation of the Number of Loop Candidates}
\label{table:loop candidates}
\resizebox{0.49\textwidth}{!}{
\begin{tabular}{c|ccccc!{\vrule width 1.5pt}ccccc} % 굵은 세로선 추가
\toprule
\multirow{2}{*}{Dataset} & \multicolumn{5}{c!{\vrule width 1.5pt}}{Ours} & \multicolumn{5}{c}{STD} \\  % 여기서 굵은 선 추가
  & 1 & 5 & 10  & 50 & 100 & 1 & 5 & 10  & 50 & 100 \\ 
 \midrule
DC04 & 0.978 & 0.993 & 0.995 & \underline{0.997} & \textbf{0.998} & 0.584 & 0.787 & 0.822 & \underline{0.859} & \textbf{0.901} \\ 
KA04 & 0.958 & 0.985 & 0.988 & \underline{0.991} & \textbf{0.993} & 0.698 & 0.833 & 0.858 & \underline{0.882} & \textbf{0.989} \\ 
BR01 & 0.914 & 0.971 & 0.980 & \underline{0.984} & \textbf{0.985} & 0.399 & 0.597 & 0.654 & \underline{0.718} & \textbf{0.794} \\ 
\bottomrule
\end{tabular}
}
\vspace{-10pt}
\end{table}

\subsection{Evaluation Results}
Fig. \ref{fig:PRcurve and FRcurve} presents the Precision-Recall (PR) and F1 Score-Recall (FR) curves, where precision, recall, and F1 score are measured at each threshold. Table \ref{table:max F1 and AUC} summarizes the Area Under the Curve (AUC) of the PR curve and the maximum F1 scores. While SC and STD perform satisfactorily compared to ISC, they struggle in more challenging scenarios (i.e., TO, BR, and RA) due to the significant presence of dynamic objects and similar spatial structures across different locations. These issues become even more pronounced in longer trajectories. In contrast, the proposed method consistently outperforms the comparison methods, even in the challenging scenes mentioned above.   

\subsection{Ablation Study} \label{sec:ablation study}
To validate the proposed keypoint extraction method (see Section. \ref{sec:Keypoint Extraction}), we compare the performance of using ARP only, RRP only, and both combined. Fig. \ref{fig:keypoint ablation}(a) and (b) show that while individual keypoints (ARP or RRP) perform well on specific sequences, the combined use of both consistently yields superior performance across all sequences. Specifically, in the BR02 sequence (in Fig. \ref{fig:keypoint ablation}(c)), RRP achieves higher performance, whereas in the KA04 sequence (Fig. \ref{fig:keypoint ablation}(d)), ARP performs better. In TO02 (Fig. \ref{fig:keypoint ablation}(e)), both keypoints perform similarly, but their combination leads to improved results. Overall results demonstrate that each keypoints has distinct strengths, and combining them through key instance segmentation (see Section. \ref{sec:key instance segmentation}) significantly enhances performance across various scenarios.

Table \ref{table:loop candidates} presents the performance comparison between the proposed method and STD as the number of loop candidates is adjusted from 1 to 100. As the number of candidates decreases, STD's performance progressively deteriorates, whereas our method consistently maintains an AUC above 0.9, even with just one candidate. The main reason for STD's performance degradation is the local ambiguity problem \cite{yin2024outram}, where similar triangle descriptors from different locations lead to incorrect matches being included in the smaller candidate set. In contrast, the robustness of the proposed method against aforementioned is due to the high reliability of the key instance set (see Section \ref{sec:Keypoint Extraction} and \ref{sec:key instance segmentation}) and the two strategies for candidate retrieval (see Section \ref{sec:candidate retrieval}).

Table \ref{table:ablation on modules} presents that incorporating instance matching (ours (w)) not only improves average precision but also slightly reduces overall search time by retaining only meaningful matching descriptors, compared to when instance matching is not used (ours (w/o)).

%---------------------------------------
%              table:ablation on modules
%---------------------------------------
\begin{table}[t]
\centering
\caption{Ablation of the Instance Matching (w: with instance matching, w/o: without instance matching)}
\label{table:ablation on modules}
\resizebox{0.4\textwidth}{!}{
\begin{tabular}{c|c|c|ccc}
\toprule
\multirow{2}{*}{Dataset} & \multirow{2}{*}{Method}  & Average & \multicolumn{3}{c}{Time [ms]} \\
& & Precision & Descriptor & Search & Total \\
 \midrule
\multirow{2}{*}{RA01} & ours (w/o) & 0.901 & 38.4 & 21.0  & 59.5  \\
&  ours (w)  & \textbf{0.939} & 38.4 & \textbf{16.4} & \textbf{55.3}  \\
 \midrule
\multirow{2}{*}{BR02} & ours (w/o)    & 0.918 & 28.3 & 19.9 & 48.3  \\
& ours (w)     & \textbf{0.947}  &  28.3 & \textbf{16.1} & \textbf{44.5} \\
 \midrule
\multirow{2}{*}{TO02} & ours (w/o)  & 0.948 & 31.6 & 23.0 & 54.7  \\
& ours (w)  & \textbf{0.977} & 31.7 & \textbf{20.4} & \textbf{51.5} \\     
\bottomrule
\end{tabular}
}
\vspace{-10pt}
\end{table}

\section{Conclusion}
In this paper, we propose a novel 3D place recognition method, RE-TRIP, which integrates reflectivity measurements with geometry measurements. We introduce key instance segmentation that leverages Absolute Reflectivity Points (ARP) and Relative Reflectivity Points (RRP) to enhance the reliability and distinctiveness of keypoints. Building on these key instances, we employ instance-level triangle descriptors to further improve the robustness of place recognition under geometric ambiguous and dynamic environment. Extensive experiments on the HELIPR and FusionPortable datasets demonstrate its superior performance over state-of-the-art methods. We believe our work provides motivation for further exploration of LiDAR's additional information, moving beyond geometry-only approaches to tackle more complex and challenging environments.

\bibliographystyle{IEEEtran}
\bibliography{ref}

% Generated by IEEEtran.bst, version: 1.14 (2015/08/26)
\begin{thebibliography}{10}
\providecommand{\url}[1]{#1}
\csname url@samestyle\endcsname
\providecommand{\newblock}{\relax}
\providecommand{\bibinfo}[2]{#2}
\providecommand{\BIBentrySTDinterwordspacing}{\spaceskip=0pt\relax}
\providecommand{\BIBentryALTinterwordstretchfactor}{4}
\providecommand{\BIBentryALTinterwordspacing}{\spaceskip=\fontdimen2\font plus
\BIBentryALTinterwordstretchfactor\fontdimen3\font minus \fontdimen4\font\relax}
\providecommand{\BIBforeignlanguage}[2]{{%
\expandafter\ifx\csname l@#1\endcsname\relax
\typeout{** WARNING: IEEEtran.bst: No hyphenation pattern has been}%
\typeout{** loaded for the language `#1'. Using the pattern for}%
\typeout{** the default language instead.}%
\else
\language=\csname l@#1\endcsname
\fi
#2}}
\providecommand{\BIBdecl}{\relax}
\BIBdecl

\bibitem{shan2020lio}
T.~Shan, B.~Englot, D.~Meyers, W.~Wang, C.~Ratti, and D.~Rus, ``Lio-sam: Tightly-coupled lidar inertial odometry via smoothing and mapping,'' in \emph{2020 IEEE/RSJ international conference on intelligent robots and systems (IROS)}.\hskip 1em plus 0.5em minus 0.4em\relax IEEE, 2020, pp. 5135--5142.

\bibitem{mur2015orb}
R.~Mur-Artal, J.~M.~M. Montiel, and J.~D. Tardos, ``Orb-slam: a versatile and accurate monocular slam system,'' \emph{IEEE transactions on robotics}, vol.~31, no.~5, pp. 1147--1163, 2015.

\bibitem{zou2024lta}
Z.~Zou, C.~Yuan, W.~Xu, H.~Li, S.~Zhou, K.~Xue, and F.~Zhang, ``Lta-om: Long-term association lidar--imu odometry and mapping,'' \emph{Journal of Field Robotics}, 2024.

\bibitem{wang2021intensity}
H.~Wang, C.~Wang, and L.~Xie, ``Intensity-slam: Intensity assisted localization and mapping for large scale environment,'' \emph{IEEE Robotics and Automation Letters}, vol.~6, no.~2, pp. 1715--1721, 2021.

\bibitem{kim20191}
G.~Kim, B.~Park, and A.~Kim, ``1-day learning, 1-year localization: Long-term lidar localization using scan context image,'' \emph{IEEE Robotics and Automation Letters}, vol.~4, no.~2, pp. 1948--1955, 2019.

\bibitem{koide2024tightly}
K.~Koide, S.~Oishi, M.~Yokozuka, and A.~Banno, ``Tightly coupled range inertial localization on a 3d prior map based on sliding window factor graph optimization,'' \emph{arXiv preprint arXiv:2402.05540}, 2024.

\bibitem{bay2008speeded}
H.~Bay, A.~Ess, T.~Tuytelaars, and L.~Van~Gool, ``Speeded-up robust features (surf),'' \emph{Computer vision and image understanding}, vol. 110, no.~3, pp. 346--359, 2008.

\bibitem{lowry2015visual}
S.~Lowry, N.~S{\"u}nderhauf, P.~Newman, J.~J. Leonard, D.~Cox, P.~Corke, and M.~J. Milford, ``Visual place recognition: A survey,'' \emph{ieee transactions on robotics}, vol.~32, no.~1, pp. 1--19, 2015.

\bibitem{rublee2011orb}
E.~Rublee, V.~Rabaud, K.~Konolige, and G.~Bradski, ``Orb: An efficient alternative to sift or surf,'' in \emph{2011 International conference on computer vision}.\hskip 1em plus 0.5em minus 0.4em\relax Ieee, 2011, pp. 2564--2571.

\bibitem{galvez2012bags}
D.~G{\'a}lvez-L{\'o}pez and J.~D. Tardos, ``Bags of binary words for fast place recognition in image sequences,'' \emph{IEEE Transactions on robotics}, vol.~28, no.~5, pp. 1188--1197, 2012.

\bibitem{steder2010robust}
B.~Steder, G.~Grisetti, and W.~Burgard, ``Robust place recognition for 3d range data based on point features,'' in \emph{2010 IEEE International Conference on Robotics and Automation}.\hskip 1em plus 0.5em minus 0.4em\relax IEEE, 2010, pp. 1400--1405.

\bibitem{rusu2010fast}
R.~B. Rusu, G.~Bradski, R.~Thibaux, and J.~Hsu, ``Fast 3d recognition and pose using the viewpoint feature histogram,'' in \emph{2010 IEEE/RSJ international conference on intelligent robots and systems}.\hskip 1em plus 0.5em minus 0.4em\relax IEEE, 2010, pp. 2155--2162.

\bibitem{gupta2024effectively}
S.~Gupta, T.~Guadagnino, B.~Mersch, I.~Vizzo, and C.~Stachniss, ``Effectively detecting loop closures using point cloud density maps,'' in \emph{Proc. of the IEEE Intl. Conf. on Robotics \& Automation (ICRA)}, 2024.

\bibitem{kim2021scan}
G.~Kim, S.~Choi, and A.~Kim, ``Scan context++: Structural place recognition robust to rotation and lateral variations in urban environments,'' \emph{IEEE Transactions on Robotics}, vol.~38, no.~3, pp. 1856--1874, 2021.

\bibitem{he2016m2dp}
L.~He, X.~Wang, and H.~Zhang, ``M2dp: A novel 3d point cloud descriptor and its application in loop closure detection,'' in \emph{2016 IEEE/RSJ International Conference on Intelligent Robots and Systems (IROS)}.\hskip 1em plus 0.5em minus 0.4em\relax IEEE, 2016, pp. 231--237.

\bibitem{yuan2024btc}
C.~Yuan, J.~Lin, Z.~Liu, H.~Wei, X.~Hong, and F.~Zhang, ``Btc: A binary and triangle combined descriptor for 3d place recognition,'' \emph{IEEE Transactions on Robotics}, 2024.

\bibitem{yuan2023std}
C.~Yuan, J.~Lin, Z.~Zou, X.~Hong, and F.~Zhang, ``Std: Stable triangle descriptor for 3d place recognition,'' in \emph{2023 IEEE international conference on robotics and automation (ICRA)}.\hskip 1em plus 0.5em minus 0.4em\relax IEEE, 2023, pp. 1897--1903.

\bibitem{jiang2019triangle}
B.~Jiang, Y.~Zhu, and M.~Liu, ``A triangle feature based map-to-map matching and loop closure for 2d graph slam,'' in \emph{2019 IEEE International Conference on Robotics and Biomimetics (ROBIO)}.\hskip 1em plus 0.5em minus 0.4em\relax IEEE, 2019, pp. 2719--2725.

\bibitem{pfreundschuh2024coin}
P.~Pfreundschuh, H.~Oleynikova, C.~Cadena, R.~Siegwart, and O.~Andersson, ``Coin-lio: Complementary intensity-augmented lidar inertial odometry,'' in \emph{2024 IEEE International Conference on Robotics and Automation (ICRA)}.\hskip 1em plus 0.5em minus 0.4em\relax IEEE, 2024, pp. 1730--1737.

\bibitem{du2023real}
W.~Du and G.~Beltrame, ``Real-time simultaneous localization and mapping with lidar intensity,'' in \emph{2023 IEEE International Conference on Robotics and Automation (ICRA)}.\hskip 1em plus 0.5em minus 0.4em\relax IEEE, 2023, pp. 4164--4170.

\bibitem{zhao2024fmcw}
M.~Zhao, J.~Wang, T.~Gao, C.~Xu, and H.~Kong, ``Fmcw-lio: A doppler lidar-inertial odometry,'' \emph{IEEE Robotics and Automation Letters}, 2024.

\bibitem{zhang2023ri}
Y.~Zhang, Y.~Tian, W.~Wang, G.~Yang, Z.~Li, F.~Jing, and M.~Tan, ``Ri-lio: reflectivity image assisted tightly-coupled lidar-inertial odometry,'' \emph{IEEE Robotics and Automation Letters}, vol.~8, no.~3, pp. 1802--1809, 2023.

\bibitem{wang2020intensity}
H.~Wang, C.~Wang, and L.~Xie, ``Intensity scan context: Coding intensity and geometry relations for loop closure detection,'' in \emph{2020 IEEE International Conference on Robotics and Automation (ICRA)}.\hskip 1em plus 0.5em minus 0.4em\relax IEEE, 2020, pp. 2095--2101.

\bibitem{lowe2004distinctive}
D.~G. Lowe, ``Distinctive image features from scale-invariant keypoints,'' \emph{International journal of computer vision}, vol.~60, pp. 91--110, 2004.

\bibitem{calonder2010brief}
M.~Calonder, V.~Lepetit, C.~Strecha, and P.~Fua, ``Brief: Binary robust independent elementary features,'' in \emph{Computer Vision--ECCV 2010: 11th European Conference on Computer Vision, Heraklion, Crete, Greece, September 5-11, 2010, Proceedings, Part IV 11}.\hskip 1em plus 0.5em minus 0.4em\relax Springer, 2010, pp. 778--792.

\bibitem{scovanner20073}
P.~Scovanner, S.~Ali, and M.~Shah, ``A 3-dimensional sift descriptor and its application to action recognition,'' in \emph{Proceedings of the 15th ACM international conference on Multimedia}, 2007, pp. 357--360.

\bibitem{salti2014shot}
S.~Salti, F.~Tombari, and L.~Di~Stefano, ``Shot: Unique signatures of histograms for surface and texture description,'' \emph{Computer Vision and Image Understanding}, vol. 125, pp. 251--264, 2014.

\bibitem{knopp2010hough}
J.~Knopp, M.~Prasad, G.~Willems, R.~Timofte, and L.~Van~Gool, ``Hough transform and 3d surf for robust three dimensional classification,'' in \emph{Computer Vision--ECCV 2010: 11th European Conference on Computer Vision, Heraklion, Crete, Greece, September 5-11, 2010, Proceedings, Part VI 11}.\hskip 1em plus 0.5em minus 0.4em\relax Springer, 2010, pp. 589--602.

\bibitem{kim2018scan}
G.~Kim and A.~Kim, ``Scan context: Egocentric spatial descriptor for place recognition within 3d point cloud map,'' in \emph{2018 IEEE/RSJ International Conference on Intelligent Robots and Systems (IROS)}.\hskip 1em plus 0.5em minus 0.4em\relax IEEE, 2018, pp. 4802--4809.

\bibitem{di2021visual}
L.~Di~Giammarino, I.~Aloise, C.~Stachniss, and G.~Grisetti, ``Visual place recognition using lidar intensity information,'' in \emph{2021 IEEE/RSJ International Conference on Intelligent Robots and Systems (IROS)}.\hskip 1em plus 0.5em minus 0.4em\relax IEEE, 2021, pp. 4382--4389.

\bibitem{shan2021robust}
T.~Shan, B.~Englot, F.~Duarte, C.~Ratti, and D.~Rus, ``Robust place recognition using an imaging lidar,'' in \emph{2021 IEEE International Conference on Robotics and Automation (ICRA)}.\hskip 1em plus 0.5em minus 0.4em\relax IEEE, 2021, pp. 5469--5475.

\bibitem{rusu20113d}
R.~B. Rusu and S.~Cousins, ``3d is here: Point cloud library (pcl),'' in \emph{2011 IEEE international conference on robotics and automation}.\hskip 1em plus 0.5em minus 0.4em\relax IEEE, 2011, pp. 1--4.

\bibitem{yin2024outram}
P.~Yin, H.~Cao, T.-M. Nguyen, S.~Yuan, S.~Zhang, K.~Liu, and L.~Xie, ``Outram: One-shot global localization via triangulated scene graph and global outlier pruning,'' in \emph{2024 IEEE International Conference on Robotics and Automation (ICRA)}.\hskip 1em plus 0.5em minus 0.4em\relax IEEE, 2024, pp. 13\,717--13\,723.

\bibitem{jung2023helipr}
M.~Jung, W.~Yang, D.~Lee, H.~Gil, G.~Kim, and A.~Kim, ``Helipr: Heterogeneous lidar dataset for inter-lidar place recognition under spatiotemporal variations,'' \emph{The International Journal of Robotics Research}, p. 02783649241242136, 2023.

\bibitem{kim2020mulran}
G.~Kim, Y.~S. Park, Y.~Cho, J.~Jeong, and A.~Kim, ``Mulran: Multimodal range dataset for urban place recognition,'' in \emph{2020 IEEE international conference on robotics and automation (ICRA)}.\hskip 1em plus 0.5em minus 0.4em\relax IEEE, 2020, pp. 6246--6253.

\bibitem{jiao2022fusionportable}
J.~Jiao, H.~Wei, T.~Hu, X.~Hu, Y.~Zhu, Z.~He, J.~Wu, J.~Yu, X.~Xie, H.~Huang \emph{et~al.}, ``Fusionportable: A multi-sensor campus-scene dataset for evaluation of localization and mapping accuracy on diverse platforms,'' in \emph{2022 IEEE/RSJ International Conference on Intelligent Robots and Systems (IROS)}.\hskip 1em plus 0.5em minus 0.4em\relax IEEE, 2022, pp. 3851--3856.

\end{thebibliography}

\end{document}